\documentclass{article} 
\usepackage{iclr2023_conference,times}

\usepackage{amsmath,amsfonts,bm}

\def\eqref#1{equation~\ref{#1}}

\def\1{\bm{1}}

\DeclareMathAlphabet{\mathsfit}{\encodingdefault}{\sfdefault}{m}{sl}
\SetMathAlphabet{\mathsfit}{bold}{\encodingdefault}{\sfdefault}{bx}{n}

\usepackage{hyperref}
\usepackage{url}

\usepackage{graphicx}
\usepackage{booktabs}
\usepackage{comment}
\usepackage{multirow}
\usepackage{subcaption}

\title{FoveaTer: Foveated Transformer for Image Classification}

\author{Aditya Jonnalagadda\\
Electrical and Computer Engineering\\
University of California\\
Santa Barbara, CA, USA \\
\texttt{aditya\_jonnalagadda@ece.ucsb.edu} \\
\And
William Yang Wang \\
Computer Science \\
University of California\\
Santa Barbara, CA, USA \\
\texttt{william@cs.ucsb.edu} \\
\AND
B. S. Manjunath \\
Electrical and Computer Engineering \\
University of California\\
Santa Barbara, CA, USA \\
\texttt{manj@ece.ucsb.edu}
\AND
Miguel P. Eckstein \\
Electrical and Computer Engineering \& Psychological and Brain Sciences \\
University of California\\
Santa Barbara, CA, USA \\
\texttt{eckstein@psych.ucsb.edu} \\
}

\iclrfinalcopy 
\begin{document}

\maketitle

\begin{abstract}
Many animals and humans process the visual field with varying spatial resolution (foveated vision) and use peripheral processing to make eye movements and point the fovea to acquire high-resolution information about objects of interest. This architecture results in computationally efficient rapid scene exploration. Recent progress in self-attention-based vision Transformers, an alternative to the traditionally convolution-reliant computer vision systems, allows global interactions between feature locations and increases robustness to adversarial attacks. However, the Transformer models do not explicitly model the foveated properties of the visual system nor the interaction between eye movements and the classification task. We propose Foveated Transformer (FoveaTer) model, which uses pooling regions and eye movements to perform object classification tasks using a Vision Transformer architecture. Using square pooling regions or biologically-inspired radial-polar pooling regions, our proposed model pools the image features from the convolution backbone and uses the pooled features as an input to transformer layers. It decides on subsequent fixation location based on the attention assigned by the Transformer to various locations from past and present fixations. The model uses a confidence threshold to stop scene exploration. It dynamically allocates more fixation/computational resources to more challenging images before making the final image category decision. We construct a Foveated model using our proposed approach and compare it against a Baseline model, which does not contain any pooling. Using five ablation studies, we evaluate the contribution of different components of the Foveated model. We perform a psychophysics scene categorization task and use the experimental data to find a suitable radial-polar pooling region combination. We also show that the Foveated model better explains the human decisions in a scene categorization task than a Baseline model. On the ImageNet dataset, the Foveated model with Dynamic-stop achieves an accuracy of $8\%$ below the Baseline model with a throughput gain of $76\%$. Using a Foveated model with Dynamic-stop and the Baseline model, the ensemble achieves an accuracy of $0.7\%$ below the Baseline using the same throughput. We demonstrate our model's robustness against PGD adversarial attacks with both types of pooling regions, where we see the Foveated model outperform the Baseline model.
\end{abstract}

\section{Introduction}

Many mammals, including humans, have evolved a locus (the fovea) in the visual sensory array with increased spatial fidelity and use head and eye movements~\citep{Land_2012, Marshall_2014} to orient such locus to regions and objects of interest. The system design allows visual-sensing organisms to accomplish two objectives: fast target detection crucial for survival and savings in computational cost. Computational savings 
are accomplished by limiting the number of units with high computational costs (i.e., higher spatial resolution processing) to the fovea's small spatial region.
Fast target detection is achieved by distributing the remaining computational power across a much larger area in the periphery, with a lower spatial resolution with increasing distance from the fovea.
Critical to the design is an efficient algorithm to guide through eye movements the high-resolution fovea to regions of interest using the low-resolution periphery~\citep{Hayhoe_2005, Strasburger_2011, Ludwig_2014} and allow optimizing the target detection and scene classification. Various computational models were proposed to model the search using foveated visual system~\citep{Yamamoto_1996, prince_2005}.

Computer vision has evolved from hand-crafted features to data-driven features in modern CNNs.
Due to their computational limitations, the objectives of the computer vision systems align well with those of human visual system: to optimize visual detection and recognition with an efficient computational and metabolic footprint.
Approaches toward saving computational power can be seen; for example, computer vision systems evolved from using sliding windows to RCNN's~\citep{R-CNN} use of selective search and Faster-RCNN's~\citep{Faster_R-CNN} use of Region Proposal Network (RPN).

A system that mimics human vision by processing the scene with a foveated system and rational eye movements has also been proposed. This approach to exploring the scene can be seen in models like RAM~\citep{RAM_model} for recognizing handwritten single-digits or detecting objects~\citep{Akbas_2017} where they sequentially process the image and decide what to process next by using the peripheral information. These foveated models approach 
that of full-resolution models but using a fraction of the computations. 
Foveated systems have also shown to result in more robustness~\citep{poggio_2015, Arturo_2020, ono_2020, poggio_2020} against adversarial attacks.

There has been a recent innovation in computer vision using Transformers~\citep{touvron2020deit, dosovitskiy2020} for object classification tasks that depart from the traditional over-reliance on convolutions.
Even after replacing the convolutions with attention modules and multilayer perceptrons, Vision Transformers~\citep{dosovitskiy2020, touvron2020deit} achieve close to state-of-the-art performance on the ImageNet dataset and provide better robustness against adversarial attacks~\citep{Shao2021OnTA}.

Due to the flattened architecture of the transformers, it is easier for multi-resolution features to share the same feature channels.
Transformers~\citep{2017_Vaswani} have the added benefit of self-attention, which facilitates the interaction of various parts of the image irrespective of distance.
No papers have evaluated the additional potential gains of incorporating a foveated architecture into Vision Transformers for the task of ImageNet classification. 

Here, we evaluate the effect of a foveated architecture and sequential eye movements on a state-of-the-art transformer architecture.
We compare the Foveated transformer relative to the Baseline model in terms of classification accuracy and robustness to adversarial attacks. 
We perform a psychophysics experiment for a scene classification task and evaluate the Foveated model agreement with the human decision against that of the Baseline model.
We first perform an object classification task using multiple fixations, moving foveal attention across different parts of the image, and using only a limited portion of the image information at each fixation, thereby reducing the input to the transformer by many folds.
The model decides on subsequent fixation location using the self-attention weights accumulated from the previous fixations until the current step. Finally, the model makes the final classification decision.





\section{Related work}

\noindent \textbf{Transformers} have achieved great success in Natural Language Processing since their introduction by~\citet{2017_Vaswani} for machine translation. 
Recently, the application of Transformer models in Computer Vision has seen tremendous success. 
Vision Transformer (ViT) model introduced by~\citet{dosovitskiy2020} achieved remarkable performance on ImageNet~\citep{ImageNet_2009} by using additional data from JFT 300M~\citep{JFT_300M} private dataset. 
Subsequently, the DeiT model~\citep{touvron2020deit} introduced knowledge transfer concepts in transformers to leverage the learning from existing models. 
Using augmentation and knowledge transfer, the DeiT model achieved close to state-of-the-art performance using training data from the ImageNet dataset alone.

\noindent \textbf{Sequential processing} provides three main advantages in computer vision.
\citet{Hinton_2010} proposed a model based on the Boltzmann machine that uses foveal glimpses and can make eye movements.
\underline{First}, it can limit the amount of information processed at a given instant to be constant, i.e., the ability to keep computations constant irrespective of the input image size.
\underline{Second}, sequential models can help model human eye movement strategies and help transfer that information to build better computer vision systems.
RAM~\citep{RAM_model} introduced a sequential model capable of making a sequence of movements across the image to integrate information before classification. In addition, the hard-attention mechanism, implemented using reinforcement learning, was used to predict the sequence of fixation locations. 
\citet{Ba_Minh_2015} extended these ideas to recognize multiple objects in the images on a dataset constructed using MNIST.
\underline{Third}, sequential processing requires fewer parameters than a model using full-resolution image input.
Other models~\citep{Xu_2015} have proposed image captioning models based on both hard-attention and soft-attention.  
Additionally, the spatial bias introduced into CNNs due to padding~\citep{alsallakh2021mind} can be overcome using sequential models~\citep{Tsotsos2011ACP}.
On the flip side, sequential models might suffer longer processing times due to sequential processing and slow convergence times for reasons similar to RNNs~\citep{Pascanu2013OnTD}.

\noindent \textbf{Computational models of categorization and eye movements} have been proposed for rapid categorization in terms of low-level properties such as spatial envelopes~\citep{Oliva_2001} and texture summary statistics~\citep{Rosenholtz_2012}.
Saliency-based models~\citep{Koch_1987, Koch_1998, Itti_2000} traditionally tried to model eye movements by identifying bottom-up properties in the image that will capture attention.
\citet{Torralba_2006} showed how saliency could be combined with contextual information to guide eye movements. 
Low-resolution periphery and high-resolution central fields are integrated with saliency to predict human-like eye movements~\citep {wloka_2018}.
Data-driven scan path prediction models~\citep{Kmmerer2022DeepGazeIM} train on image content and human fixations to predict the fixations under a free viewing but do not consider decision accuracy in specific tasks after multiple fixations.
Goal-directed attention control~\citep{Zelinsky2021PredictingGA} showed the dependency of search patterns on target features and scene context.
\citet{Akbas_2017} implemented a biologically-inspired foveated architecture~\citep{Metamers_2011} with a deformable parts model to build a foveated object detector on PASCAL dataset~\citep{Everingham2014ThePV}, whose accuracy was close to a full-resolution model but using a fraction of the computations.
Spatial transformer networks~\citep{ST_2015}, an older technique different from the proposed Vision Transfomers, were used on CIFAR-10 dataset~\citep{Krizhevsky2009LearningML}, with foveation to improve object localization using foveated convolutions~\citep{Harris_2019} and achieve better eccentricity performance~\citep{Dabane_2021} on MNIST dataset~\citep{LeCun1998GradientbasedLA}.

\noindent \textbf{FoveaTer} combines biologically-inspired foveated architecture with a Vision Transformer Network.
Unlike the previous architectures~\citep{Akbas_2017,Mnih2014}, we do not scale the image and thereby retain the parallelism with biological mechanisms.
We apply our model to real-world images from the ImageNet dataset for image classification. In contrast, the previous works were mainly limited to datasets with small image sizes or a smaller number of output classes. They did not extend to large-scale real-world databases like ImageNet, which has $1000$ class labels. We also evaluate the functional roles of various components through ablation studies, including the memory of foveal and peripheral information from previous fixations,  inhibition of return,  and eye movement guidance algorithms.

A novel aspect of the proposed work is that the model also learns that all images are not equally difficult to classify, adapting the exploration of eye movements to different images and thus varying computational resources used to classify different images successfully.
The model implements this idea using a confidence threshold to restrict the scene exploration to the necessary fixations to classify the image.

Also novel is an evaluation of the adversarial robustness of our model to understand the contributions of the foveated architecture and that of sequential fixations towards defense against adversarial attacks.
We use the projected gradient descent method~\citep{PGD_2016, Mardy_2019}, which iteratively computes the adversarial image.
The architectural changes may not be trivially transferable to a new architecture. End-to-End training and hyper-parameter settings might be needed to adapt to the architectural differences.

\section{Model}

\begin{figure}[ht]
    \vspace{-1em}
     \centering
     \begin{subfigure}[b]{0.8\textwidth}
         \centering
         \includegraphics[width=\textwidth]{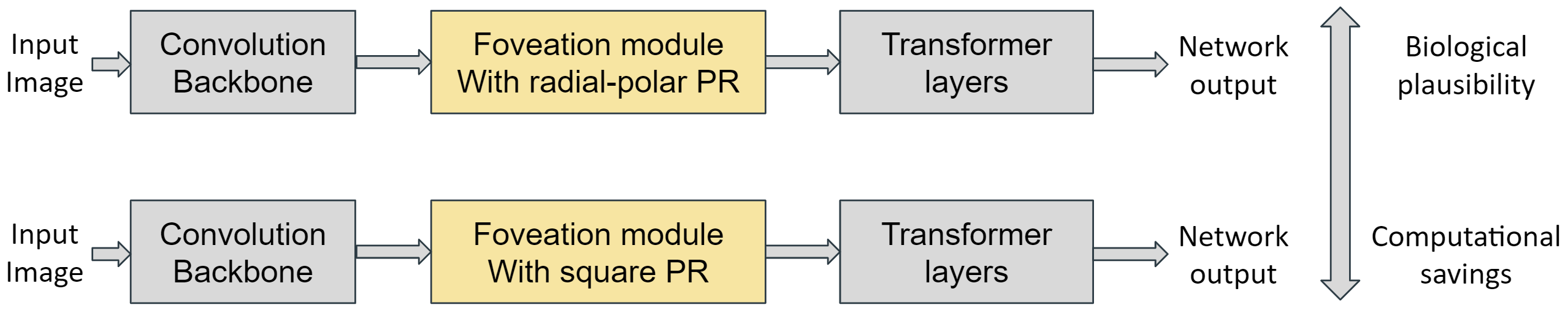}
     \end{subfigure}
    \caption{PR refers to the pooling regions. Foveation with radial-polar pooling regions is more biologically plausible than the square pooling regions but computationally slower and vice-versa.}
    \label{fig:Different_models}
\end{figure}

The model consists of three components, as shown in Figure~\ref{fig:Different_models} - convolution backbone, foveation module, and transformer layers.
Interactions between different feature locations are limited to local regions in the convolution backbone.
The Foveation module performs non-uniform pooling on the input features, reducing feature dimensionality.
The Foveation module can contain two types of pooling regions, square pooling regions which provide computationally fast processing, or biologically plausible radial-polar pooling regions, ~\citep{Metamers_2011}.
Under this non-uniform average-pooling model, locations closer to the fixation location use smaller neighborhoods for pooling than locations far from the fixation location.
The last component consisting of the transformer layers contributes in three ways - 1. They allow global interactions, which allows the possibility of using context-based decision-making. 2. They eliminate the need to design convolution layers on top of non-uniform sampled features from the Foveation module. 3. Self-attention weights of the transformer layers can be helpful in fixation guidance.

\begin{figure}[t]
    \vspace{-1em}
     \centering
     \begin{minipage}[b]{0.65\textwidth}
         \begin{subfigure}[b]{1.0\textwidth}
             \centering
             \includegraphics[width=0.55\textwidth]{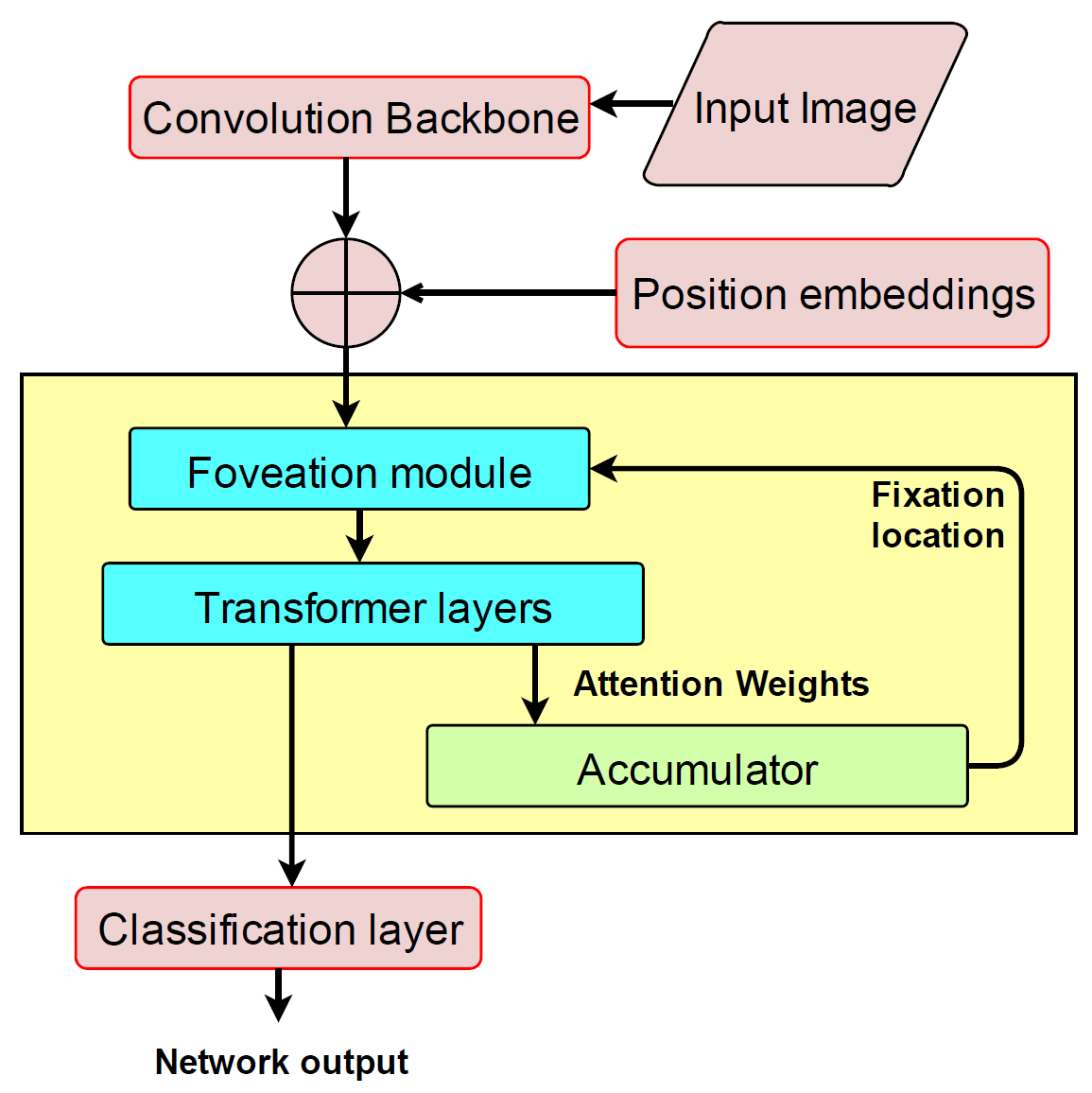}
             \caption{\noindent \textbf{FoveaTer architecture: }The foveation module performs fixation-dependent pooling. \textit{Accumulator} uses the attention weights from the last transformer layer of past and present fixations to predict the next fixation location. Model blocks within the yellow region are executed for each fixation.}
             \label{fig:FoveaTer_Architecture}
         \end{subfigure}
     \end{minipage}
     \hfill
     \begin{minipage}[b]{0.3\textwidth}
         \begin{subfigure}[b]{1.0\textwidth}
             \centering
             \includegraphics[width=0.75\textwidth]{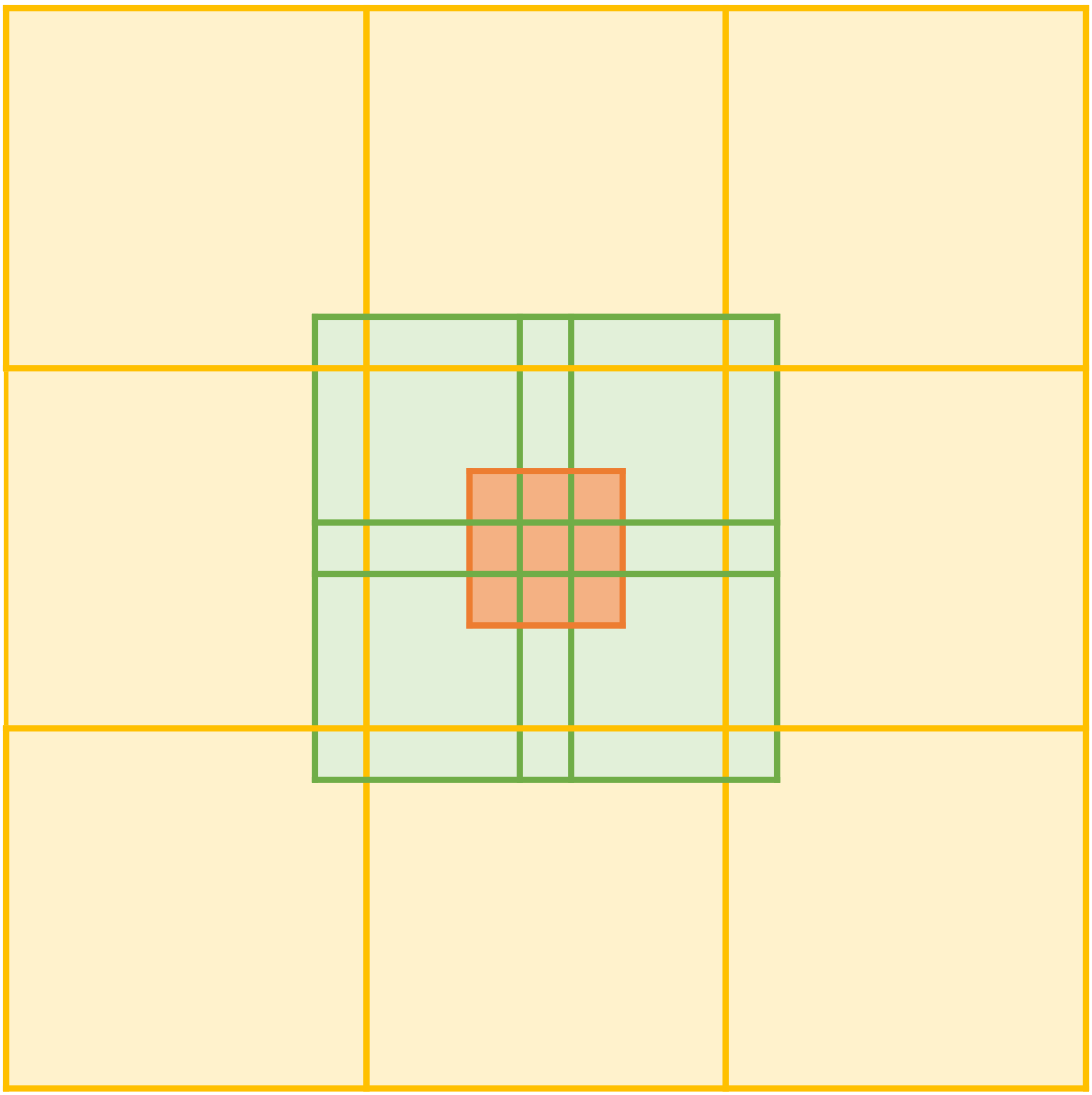}
             \caption{\noindent \textbf{Square pooling:} Input feature map is pooled to generate the pooled feature vectors. The Fovea is shown in red containing $9$ feature vectors. The first level of pooling regions is of size $5\times5$ with stride $4$ (green). The second pooling region level is size $7\times7$ with stride $7$ (orange).}
             \label{fig:square_PR}
         \end{subfigure}
     \end{minipage}
     \caption{Network architecture}
     \vspace{-1em}
\end{figure}

For the square pooling regions, the input image is first passed through the convolution backbone resulting in a feature vector of size $[384, 14, 14]$.
After adding the sinusoidal position embedding and performing fixation-dependent average-pooling using the Foveation module, the feature size reduces to $[384, 22]$.
Pooled features of size $[384, 22]$ are passed through the transformer layers, followed by the classification layer resulting in a logits vector.
We use the self-attention weights from the last transformer layer to predict the subsequent fixation location.
We make five fixations on each image during the model training.
This choice keeps the computational cost relatively the same as the Baseline model.
Model architecture is shown in Figure~\ref{fig:FoveaTer_Architecture}.

The convolution backbone consists of six convolution layers and is structured similarly to the initial layers of the ResNet-18 model.
Square pooling regions can exploit the fast average-pooling library functions, whereas the pooling in the radial-polar pooling regions needs custom implementation.
Four architectural changes make it possible for the FoveaTer model to perform serial processing, achieve throughput improvements and retain information across fixations.
Firstly, the Foveation module is a plug-in module that can be preceded or succeeded by the transformer layers. However, additional changes would be required for the convolution layers to follow the foveation module.
Secondly, the periphery (i.e., the pooling regions other than Fovea) pools the feature vectors and, as a result, reduces the number of features processed by the subsequent layers.
Thirdly, the attention-based fixation guidance mechanism (FGM) helps predict the subsequent fixation location using the attention values of current and past fixations.
Lastly, the features from the past fixation's foveal locations are retained and processed along with the foveal and peripheral features of the current fixation. Thus, allowing the model to access memory.

\textbf{Retention of foveal features: }
The number of feature vectors processed by the Foveation module varies across fixations due to the retention of foveal features from past fixations. For each fixation, if the number of peripheral and foveal features are A and B, the number of features processed by the foveation module at Nth fixation equals A+NB.

\textbf{Initial fixation for the Foveated model:}
The input feature map to the Foveation module has a spatial size of $14\times14$ for the condition of square pooling regions.
All locations except the last and first row/column are potential fixation locations, resulting in $144$ locations.
We select a random location as the initial fixation point during training, and the model guides subsequent fixations.

\textbf{Loss function: }We use Cross-entropy for computing the classification loss. 
Loss from all fixations is incorporated to get the mini-batch loss, $loss =   \sum_{i=1}^{N} L_{CE}(O_{i}, y)$ Where $i$ corresponds to the fixation index, $N=5$ for the Foveated model \& $N=1$ for the Baseline model, i.e., single-pass, $y$ corresponds to the target label, $O_{i}$ corresponds to the network output for fixation $i$, and $L_{CE}$ correspond to cross-entropy loss.

\subsection{Foveation module}
\label{Foveation_module}

The mean feature vector corresponding to each pooling region is computed using $P = (1/M) \sum_{j=0}^{M-1}E_{j}$, Where $E_{j}$ is a feature vector belonging to that pooling region, and $M$ is the number of feature vectors in that pooling region.

We use square pooling regions for computational speed-up.
Each image in a mini-batch has a corresponding fixation location. The fixation location represents the center of the visual field, allowing us to align the input image/feature map with the visual field.
After aligning the input feature map with the visual field, features falling within a pooling region are average-pooled, and the resultant pooled vector represents that pooling region.
We use pooling regions with receptive field sizes $1\times1$, $5\times5$, $7\times7$ blocks on a feature map of size $14\times14$ blocks, as shown in Figure~\ref{fig:square_PR}.
Each block corresponds to a $[16,16]$ pixel region in the input image of width and height $224$.
Central $3\times3$ red block represents the high-resolution Fovea, where there is no average pooling.
The next ring of pooling regions, where the pooling region is green, has a receptive field of $5\times5$ which translates to an average pooling of 25 feature vectors to generate the representative feature vector for that pooling region.
Similarly, the rings of orange-colored pooling centers have receptive fields of $7\times7$.

\subsection{Accumulator}
Accumulator uses the self-attention weights from the last transformer layer for fixation guidance.
Using the attention weights of the current and past fixations along with inhibition of return, the Accumulator (see below) predicts the subsequent fixation location.
A confidence map ($CM_{N}$) is constructed based on the fixation point location by putting these weights back on a $14\times14$ map at the corresponding pooling region's location, where $14\times14$ corresponds to the size of the input feature map.
Inhibition of return (IoR)~\citep{Dukewich2015} refers to a tendency in human observers not to attend to previously attended or fixated regions.
Old accumulated attention map ($AM_{N-1}$) is weighted by $0.5$ and added to the current confidence map to create the new accumulated attention map: $AM_{N} = 0.5*AM_{N-1} + CM_{N}$.
The inhibition of return ($IOR_{N}$) map is initialized with zeros and is the same size as the feature map. Locations corresponding to the current fovea location are changed to $16$.
After subtracting the IOR map from the accumulated attention map, max location of the resultant map is used as the next fixation location $Fix_{N+1} = \arg\max{(AM_{N} - IOR_{N})}$.

\subsection{Dynamic-stop of Fixation Exploration:}
Due to various factors such as occlusion, camera angle, and brightness, the difficulty of making a classification decision varies across object classes and images. 
To achieve higher computational efficiency in our Foveated model during inference, we stop exploring the images with fixations when the predicted class with the highest probability reaches a pre-defined threshold corresponding to that class.
We compute the threshold from the training dataset's set of all the correct prediction probabilities.
The model stops if the top prediction is above the $50$th percentile of probabilities for that class and the second-best prediction is below the $5$th percentile for that respective class.

\section{Ablation Studies}

\begin{table}
    \vspace{-3em}
	\begin{minipage}{0.55\linewidth}
		\caption{\textbf{Ablation Studies:} Four network components are considered, and the percentage accuracy drop after five fixations with respect to the Benchmark model is reported in the last row. Checkmark (\checkmark) indicates that the model includes the component, while the dashed-line (---) indicates that the component has been removed.}
		\label{Ablation_1}
		\centering
		\resizebox{1.0\columnwidth}{!}{
        \begin{tabular}{c|cccccc}
            \toprule
            \multicolumn{1}{c}{\bf Network component} &\multicolumn{1}{c}{\bf Benchmark} &\multicolumn{1}{c}{\bf Study 1} 
            &\multicolumn{1}{c}{\bf Study 2} &\multicolumn{1}{c}{\bf Study 3}
            &\multicolumn{1}{c}{\bf Study 4}\\
            \midrule
            Foveation  & \checkmark & \checkmark & \checkmark & \checkmark & \checkmark\\
            Peripheral features  & \checkmark & ---  & \checkmark & \checkmark & \checkmark\\
            Foveal features  & \checkmark  & \checkmark & --- & \checkmark & \checkmark\\
            Retention of foveal features   & \checkmark & \checkmark  & ---  &  ---        & \checkmark \\
            Inhibition of return   & \checkmark & \checkmark & \checkmark  & \checkmark &  --- \\
            \midrule
            Accuracy@1      &  $76.29$  &  $62.85$  & $72.60$   & $75.29$ & $75.23$\\
            Percentage drop &     & $17.6$  & $4.8$ &  $1.3$ & $1.4$\\ 
            \bottomrule
        \end{tabular}}
	\end{minipage}\hfill
	\begin{minipage}{0.4\linewidth}
		\centering
		\includegraphics[width=\textwidth]{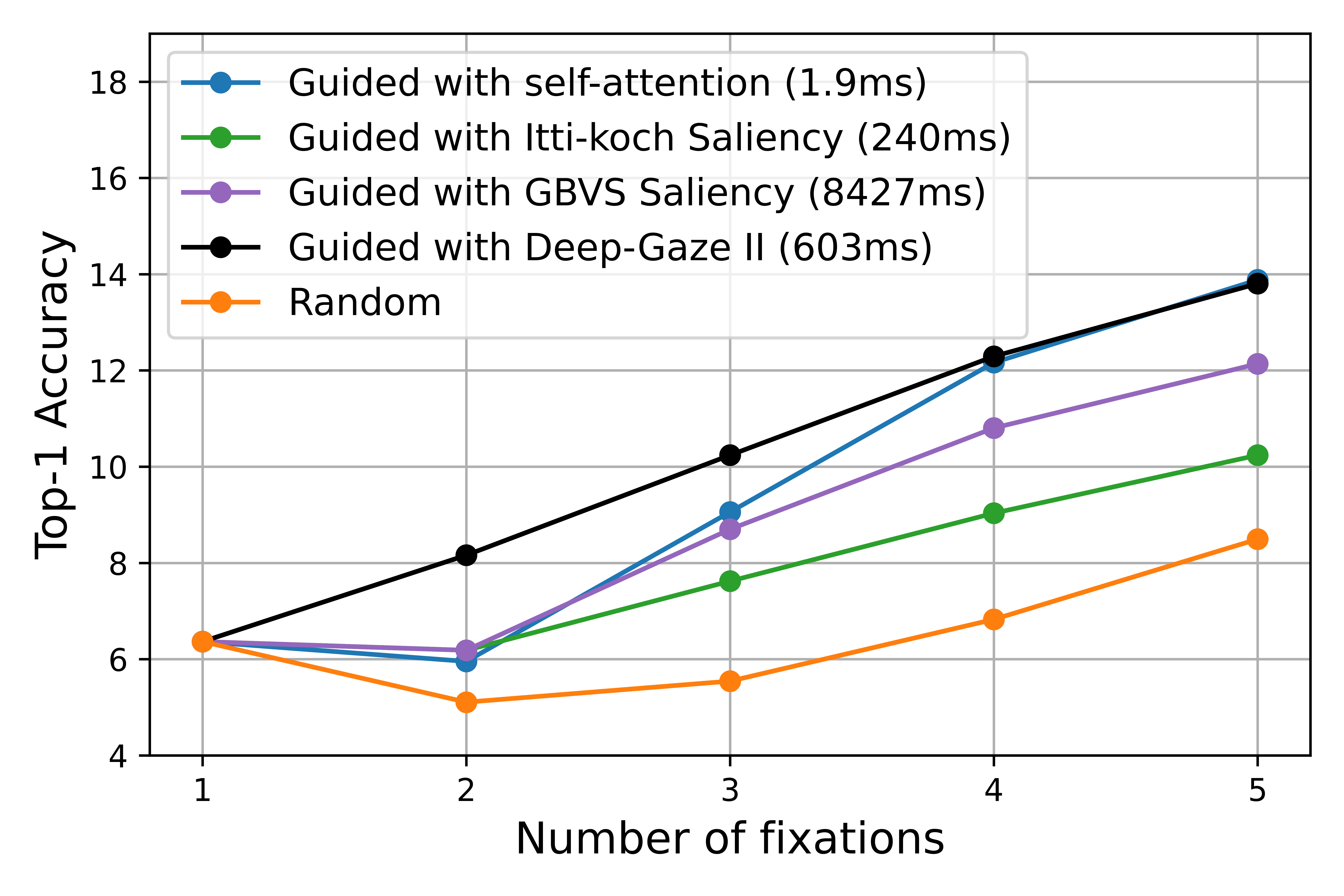}
        \captionof{figure}{\textbf{Study 5: }Self-attention guidance outperforms the random fixations by $63\%$. Initial fixation at the top-left corner. Time taken for computing five fixations is shown in brackets.}
		\label{fig:Rand_Guided}
	\end{minipage}
	\vspace{-1em}
\end{table}

We study the contribution of different network components to model performance in five ablation studies.
The model was trained on ImageNet for $300$ epochs and fine-tuned for $30$ epochs for each ablation study.
The first two studies asses the importance of peripheral and foveal features.
Studies three and four assess the importance of memory provided by past foveal features and IoR, respectively.
Lastly, we look at the contributions of the fixation guidance mechanism.
Results are shown in Table~\ref{Ablation_1} and Figure~\ref{fig:Rand_Guided}. 

\textbf{Study 1: Contribution of peripheral features: }
Peripheral features are essential because they contribute to image classification and help decide the subsequent fixation location.
There is a sharp $17.6\%$ drop in the network performance by removing the peripheral features.

\textbf{Study 2: Contribution of foveal features: }
Foveal features provide high-resolution information. By removing access to foveal features of current and past fixations, the model loses access to all full-resolution information.
There is a $4.8\%$ drop in the network performance by removing the foveal features.

\textbf{Study 3: Retention of the foveal features: }
We incorporate memory by retaining the past foveal features and processing them along with the foveal and peripheral features of the current fixation.
Even without this network component, the network has some memory as the model makes fixations to more informative locations using guided fixations.
In this experiment, we remove the usage of foveal features from past fixations, and as a result, the model performance drops by $1.3\%$.

\textbf{Study 4: Contribution of Inhibition of Return: }
By limiting the model's ability to revisit the fixation locations of the past, we force the model to explore rather than get stuck at one location. 
We only see a slight drop in performance of $1.4\%$ without the IoR, signifying that the model can operate well without IoR. The results suggest that the model can learn not to revisit locations without explicitly implementing IoR.

\textbf{Study 5: Effectiveness of Fixation guidance mechanism: }Objects in the ImageNet dataset often occupy a large part of the image. As a result, image classification might be possible by fixating anywhere on a large percentage of the image. The importance of guided fixation is best illustrated when a few image regions are informative. To identify that subset of images, we separate the testing images into two groups, one with moderate difficulty and the other with too few or too many informative locations. To identify these two groups of images, we run our model under a one-fixation condition at each possible fixation location and calculate the percentage of locations (PoL) with the correct classification in that image.
We use this as the metric for image difficulty, i.e., higher PoL signifies less difficulty and vice versa. 
As there are $144$ locations, PoL ranges from $0$ to $144$. We label all the images where the PoL is more than one-eighth the maximum value, i.e., greater than $18$, as too easy. Similarly, images with a PoL of zero are labeled as too difficult. After removing the images labeled as too easy or too difficult, approximately $8\%$ images fall in the middle, i.e., moderately difficult category. 
Figure~\ref{fig:Rand_Guided} shows the comparison of random and guided fixations on this subset of images, and guided fixations have approximately $63\%$ improvement over random fixations.
We also compare the fixation guidance using the Itti-Koch, Graph Based Visual Saliency~\citep{Harel2006GraphBasedVS} and the DeepGaze-II model, where they take the image without foveation as input.
Fixations guided by self-attention outperform the fixations guided by the Itti-Koch model and are as effective as those guided by the Deep-Gaze II for later fixations.  Lower performance than Deep-Gaze II in the first fixations is not surprising since Deep-Gaze II is predicting the most likely regions to be fixated by humans (not the order).

\begin{figure}[t]
    \vspace{-3em}
    \begin{minipage}[b]{0.66\linewidth}
        \centering
        \includegraphics[width=0.7\textwidth]{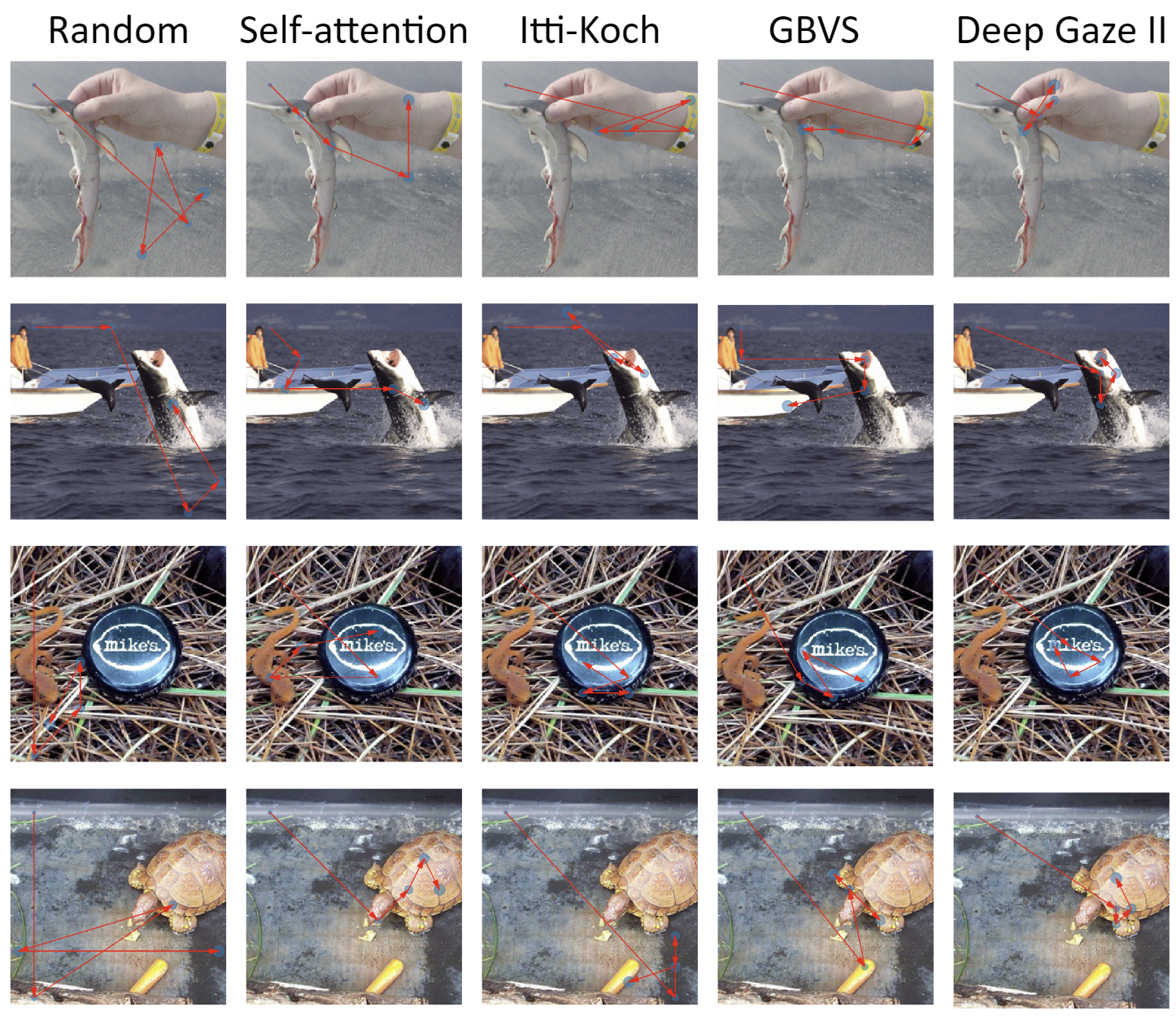}
        \caption{Fixation guidance by different models}
         \label{fig:Model_fixations}
    \end{minipage}
    \hspace{0.1cm}
    \begin{minipage}[b]{0.3\linewidth}
        \centering
        \includegraphics[width=\textwidth]{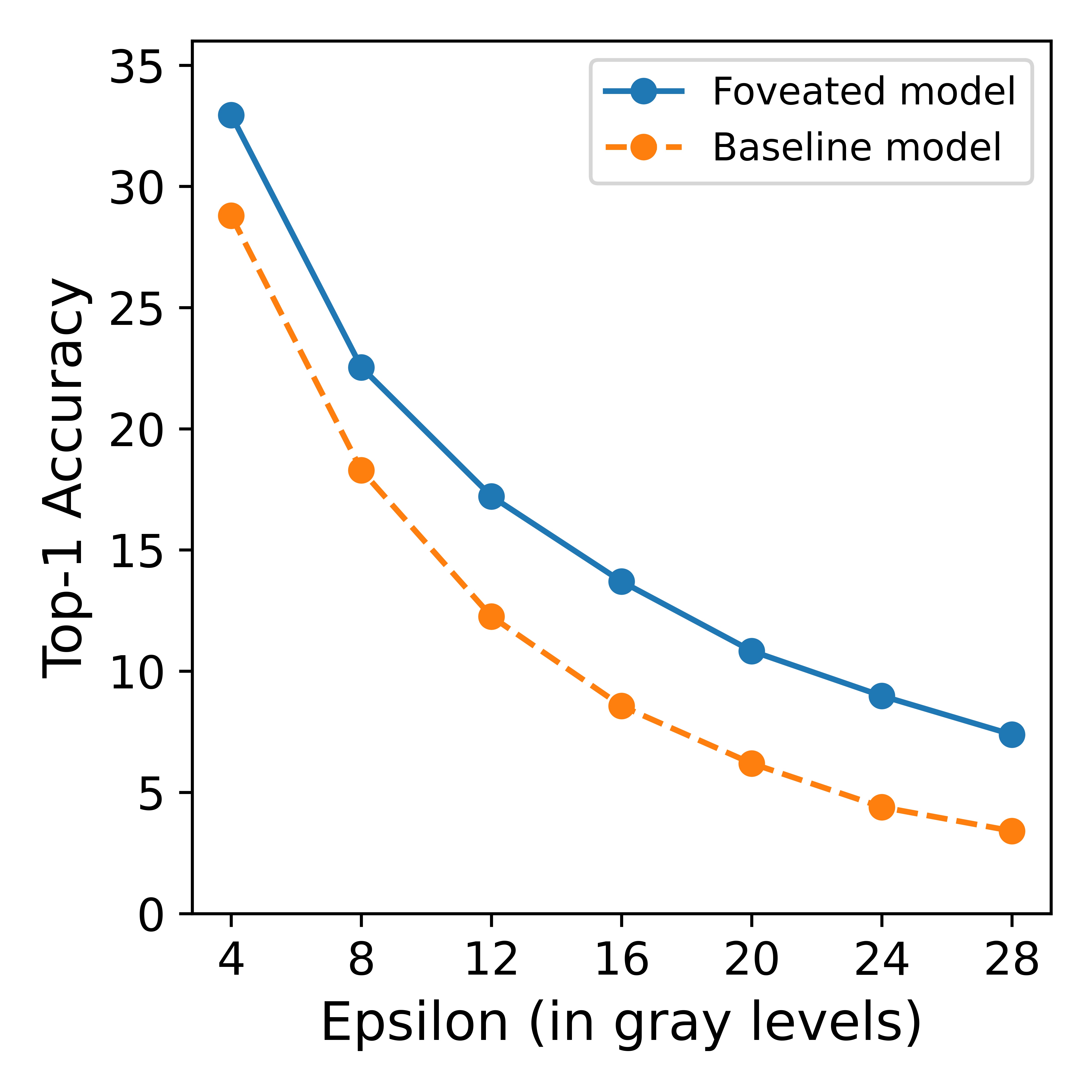}
        \caption{PGD attack: Strength of the attack is represented in terms of equivalent gray levels.}
        \label{fig:Adversarial_attack}
    \end{minipage}
    \vspace{-1em}
\end{figure}

Comparing the time taken for fixation prediction, our fixation guidance is the fastest as we leverage the model's internal attention weights rather than running a separate model.
Time taken for computing five fixations - Guided by self-attention ($1.9$ms) $<$ Itti-Koch ($240$ms) $<$ Deep-Gaze ($603$ms) $<$ GBVS ($8427$ms).
Sample image fixations are shown in Figure~\ref{fig:Model_fixations}.



\section{Accuracy and Robustness on ImageNet}

In the following sub-sections, we compare the performance, computational complexity, and adversarial robustness of the Foveated model against the Baseline. 
The Foveated model is trained for five fixations, although it can work with any desired number of fixations at test time.
Baseline and Foveated models have the same $24$M parameters.

We use the Patchconvnet~\citep{touvron2021patchconvnet} architecture.
We initialize the convolution backbone with the weights from ResNet-18~\citep{He2016DeepRL} model, and the transformer layers are initialized with the weights of the DeiT-small~\citep{touvron2020deit} model and trained for $300$ epochs with an initial learning rate of $5e-4$ and a minimum learning rate of $1e-5$.
We use AdamW~\citep{Adam_2014, AdamW_2019} optimizer with a decay of $1e-8$ and a cosine learning rate schedule.
We use ImageNet~\citep{ImageNet_2009} dataset for the results shown in the following sub-sections.
We use RTX A6000 GPUs for training and testing purposes.
We report the number of inferences completed by the GPU during a one-second time interval to compare the computational complexity of different models during inference time.

\subsection{Top-1 Accuracy:}

For the Dynamic-stop, we first compute the throughputs of the Foveated model for each of the one to five fixation conditions, followed by the number of images belonging to each of those five fixation conditions. 
The throughput of the Dynamic-stop model is computed as the weighted Harmonic mean of the throughputs of individual fixation models.
Ensemble refers to a model composed of both the Foveated and Baseline models.
When the Dynamic-stop is applied, and the model cannot make a decision even after the maximum number of fixations, the Ensemble model transfers the responsibility of making a decision to the Baseline model.

We present the results on the ImageNet dataset in Table~\ref{ImageNet_results_table}.
The Deit-Small model has a throughput of $1699$ and Top-1 accuracy of $79.83$. 
The Baseline, which has the same architecture as the Foveated model except for the foveation module, has a throughput of $1229$ and an accuracy of $81.90$.
Since the first level of the pooling region is of size $5\times5$, we construct a pooled version of the baseline model using $5\times5$ average-pooling.
We compare this with the Foveated model with two fixations, with approximately the same throughput.
The Foveated model with two fixations outperforms the uniformly pooled Baseline model, as shown in row $6$.
Dynamic-stop and Ensemble performances are shown in the last two rows.
The performance of the ensemble model reaches close to the Baseline model in terms of throughput and accuracy.

\begin{table}
  \vspace{-3em}
  \caption{Throughput and Accuracy on ImageNet: We compare our models against the baseline model using Top-1 accuracy and Image throughput. (\textit{Uniform pool} - uniform $5\times5$ pooling, \textit{CF} - initial fixation at image center, \textit{Rand} - random initial fixation)}
  \label{ImageNet_Results_table}
  \centering
  \resizebox{0.65\columnwidth}{!}{
  \begin{tabular}{c|c|cc|cc}
    \toprule
    \multicolumn{1}{c}{Model} & \multicolumn{1}{c}{Pooling type} & \multicolumn{1}{c}{Fixations} & \multicolumn{1}{c}{Type} & \multicolumn{1}{c}{Throughput} & \multicolumn{1}{c}{Acc@1} \\
    \midrule
    DeiT-Small &    &         &    &     $1699$    &   $79.83$           \\
    \midrule
      \multirow{2}{4em}{Baseline} &  &         &    &    $1229$ &          $81.90$    \\
      &   &         &  Uniform pool  &     $2506$    &   $70.90$  \\
    
    
       
    \midrule
     \multirow{7}{4em}{Foveated} &  \multirow{7}{4em}{Square} &    Rand-1     &    &      $3820$   &  $69.80$   \\
     \cmidrule(r){3-6}
     &  &    CF-1     &     &    $3820$    &    $72.80$    \\
     &  &    CF-2     &     &    $2307$    &    $74.70$    \\
     &  &    CF-3     &     &    $1506$    &    $75.40$    \\
     &  &    CF-5     &     &    $923$    &    $76.30$    \\
     \cmidrule(r){3-6}
      &       & CF-3         &  Dynamic Stop         &  $2169$      &    $75.30$   \\
       &      & CF-3         &  Ensemble   &  $1236$  &    $81.30$    \\
        
    \bottomrule
  \end{tabular}}
  \label{ImageNet_results_table}
\end{table}

\subsection{Robustness against adversarial attacks}

We consider the Projected Gradient Descent (PGD) attack to compare the robustness of Foveated and Baseline models.
PGD uses ten iterations with a step-size of $\epsilon/5$ and l-infinity norm.
We use Cleverhans library~\citep{cleverhans_library} for implementing the adversarial attacks.
Figure~\ref{fig:Adversarial_attack} shows the model accuracy after attacking the input image with the adversarial attack.
Epsilon ($\epsilon$) represents the strength of the attack. 
Foveated model displays strong defense as compared to the Baseline model.
Foveated model consistently outperforms the Baseline model.
A comparison with existing models, showing the robustness of the foveated systems against adversarial attacks, is demonstrated in Appendix~\ref{comparison_with_existing_models}.

\section{Biologically plausible FoveaTer}
\begin{figure}
     \centering
     \begin{subfigure}[b]{0.8\textwidth}
         \centering
         \includegraphics[width=\textwidth]{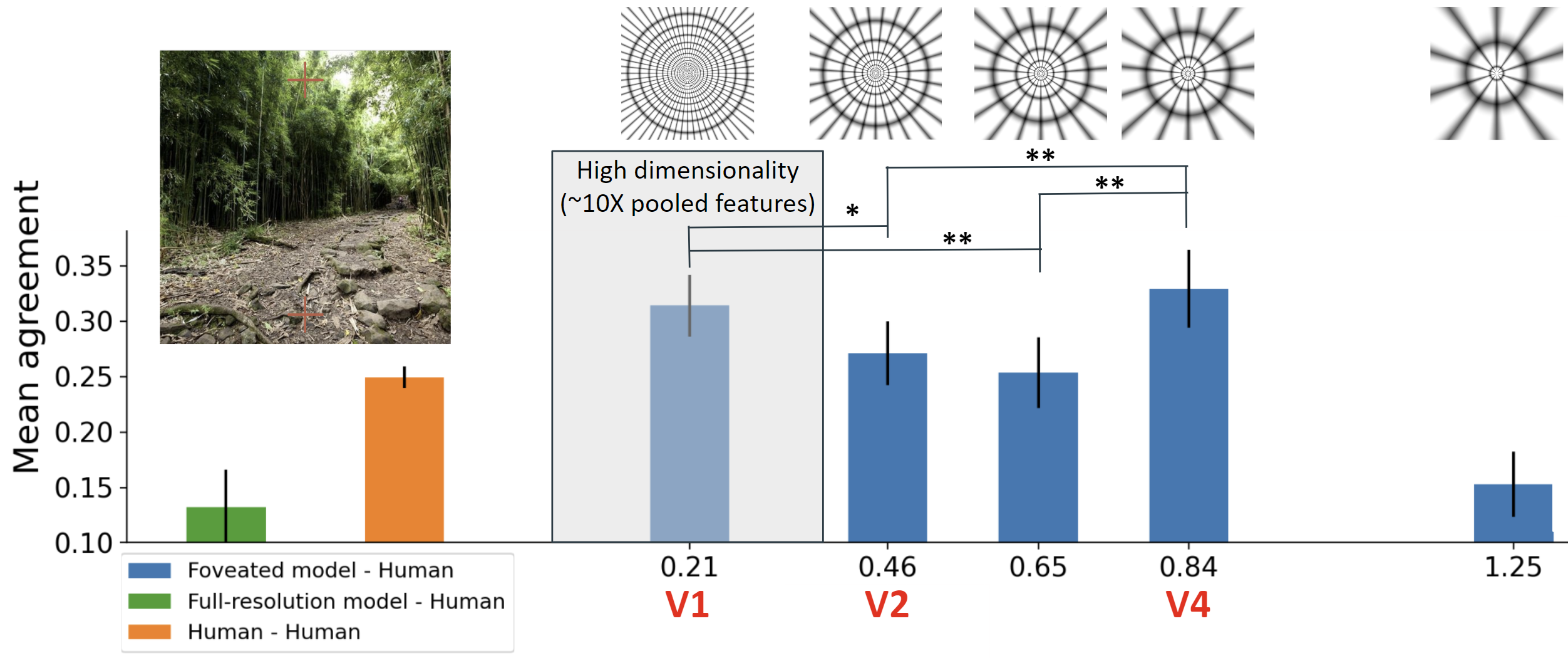}
     \end{subfigure}
     \caption{Mean agreement values of the Baseline and the Foveated models with human decisions (correct/incorrect). Error bars refer to the standard error across $22$ participants. Paired t-test p values indicate statistical significant agreement differences across scales, **$p<0.01$, *$p<0.05$.}
     \vspace{-1em}
     \label{fig:Scene_two_fixations}
\end{figure}

Radial-Polar pooling makes the model more biologically plausible.
Through psychophysics experiments of image discrimination and modeling, \citet{Metamers_2011} showed that different layers of the visual cortex correspond to different \textit{scales} where the \textit{scale} parameter determines how many radial and polar pooling regions are present in that configuration.
We use this model to predict human decisions in a scene classification psychophysics task while maintaining fixation and calibrate the scaling parameters of the pooling regions of FoveaTer.
Figure~\ref{fig:Scene_two_fixations} shows examples of various configurations.

\subsection{Calibration of radial-polar pooling regions}

We used thirty scene categories from the places365 dataset~\citep{Zhou2018PlacesA1} to create the experiment dataset.
The task was to classify each image into one of the 30 categories. Sixty images were presented, with each image subtending 22.7 x 22.7 degrees visual angle, and observers fixated at the bottom-center or top-center within the images (2.2 degrees from the top or bottom edges of the image, Figure~\ref{fig:Scene_two_fixations}). Real-time infra-red video eye tracking allowed for interruption of the displayed image when observers made an eye movement.

We tweaked the last convolutional layer so that the convolution backbone of the model outputs a $56\times56\times384$ feature map instead of a $14\times14\times384$ feature map, thus allowing us to apply the pooling regions on a higher resolution feature map.
We train multiple models with different \textit{scale} values for spatial pooling.
For each \textit{scale}, Foveated model is trained for $60$ epochs after initializing the weights with the square-pooled Foveated model trained on ImageNet. 

Error consistency metric~\citep{Geirhos2020BeyondAQ} produces the normalized decision agreement between two observers, where the normalization is a function of the accuracy of both observers.
We computed the mean agreement between the human decisions and the Foveated model for a set of \textit{scales} as shown in Figure~\ref{fig:Scene_two_fixations}.
We also computed the mean agreement between human decisions and the Baseline model (independent of \textit{scale}).
For \textit{scales} corresponding to V2 (\textit{scale}-0.46) and V4 (\textit{scale}-0.84) layers of the visual cortex, we observe a significant difference between the mean agreement of humans with Foveated and Baseline models.
Although the accuracy of the Baseline model ($0.93$) is higher than the Foveated model ($0.86$), human decisions with a mean accuracy of $0.83$ are in better agreement with the Foveated model.
The fixation at the top-center or the bottom-center limited the image information accessible to the human observers, which the Full-resolution fails to model.
Our findings suggest that human categorization of scenes within a single fixation can be better predicted with FoveaTer with pooling regions that scale according to properties of the visual cortex (V1 and V4).

\subsection{Accuracy and robustness on ImageNet}

We evaluated FoveaTer's accuracy and robustness using pooling parameters (scaling 0.84, V4) that predicted human scene classification decisions and were computationally efficient (relative to V1). Results are shown in Table~\ref{ImageNet_results_table_rad_polar}.
The throughput of the Foveated model with radial-polar pooling regions is very low due to the lack of library functions implementing the radial-polar pooling.
As the specialized hardware performing neuro-foveal pooling becomes available in the future, throughput gaps will disappear, and the Foveated model will become competitive with the Baseline models.
Adversarial robustness of Foveated model against PGD attack with radial-polar pooling regions is illustrated in Figure~\ref{fig:Adversarial_attack_rad_polar}.
As with the square pooling regions, Foveated model with radial-polar pooling regions is also more adversarial robust than the Baseline model.

\begin{table}
    \vspace{-3em}
	\begin{minipage}{0.66\linewidth}
	 \caption{Throughput and Accuracy on ImageNet using radial-polar pooling regions with \textit{Scale} $0.84$: All foveated models made three fixations. (\textit{CF} - initial fixation at image center)}
	\label{ImageNet_results_table_rad_polar}
		\centering
  \resizebox{0.7\columnwidth}{!}{
  \begin{tabular}{c|c|cc}
    \toprule
    \multicolumn{1}{c}{Model} & \multicolumn{1}{c}{Type} & \multicolumn{1}{c}{Throughput} & \multicolumn{1}{c}{Acc@1} \\
    \midrule
      \multirow{1}{4em}{Baseline}       &    &    $1229$ &          $81.90$    \\

    \midrule
     \multirow{3}{4em}{Foveated (square)} &     &    $1506$    &    $75.40$    \\
      &       Dynamic Stop         &  $2169$      &    $75.30$   \\
       &     Ensemble   &  $1236$  &    $81.30$    \\
         
         \midrule
     \multirow{3}{4em}{Foveated (Radial-Polar)}      &   & $117$ &  $76.69$   \\
    &     Dynamic Stop         &    $198$   &   $76.65$   \\
    &     Ensemble   & $186$  & $81.52$     \\

          
    \bottomrule
  \end{tabular}}
	\end{minipage}\hfill
	\begin{minipage}{0.3\linewidth}
    		\centering
    		\includegraphics[width=0.8\textwidth]{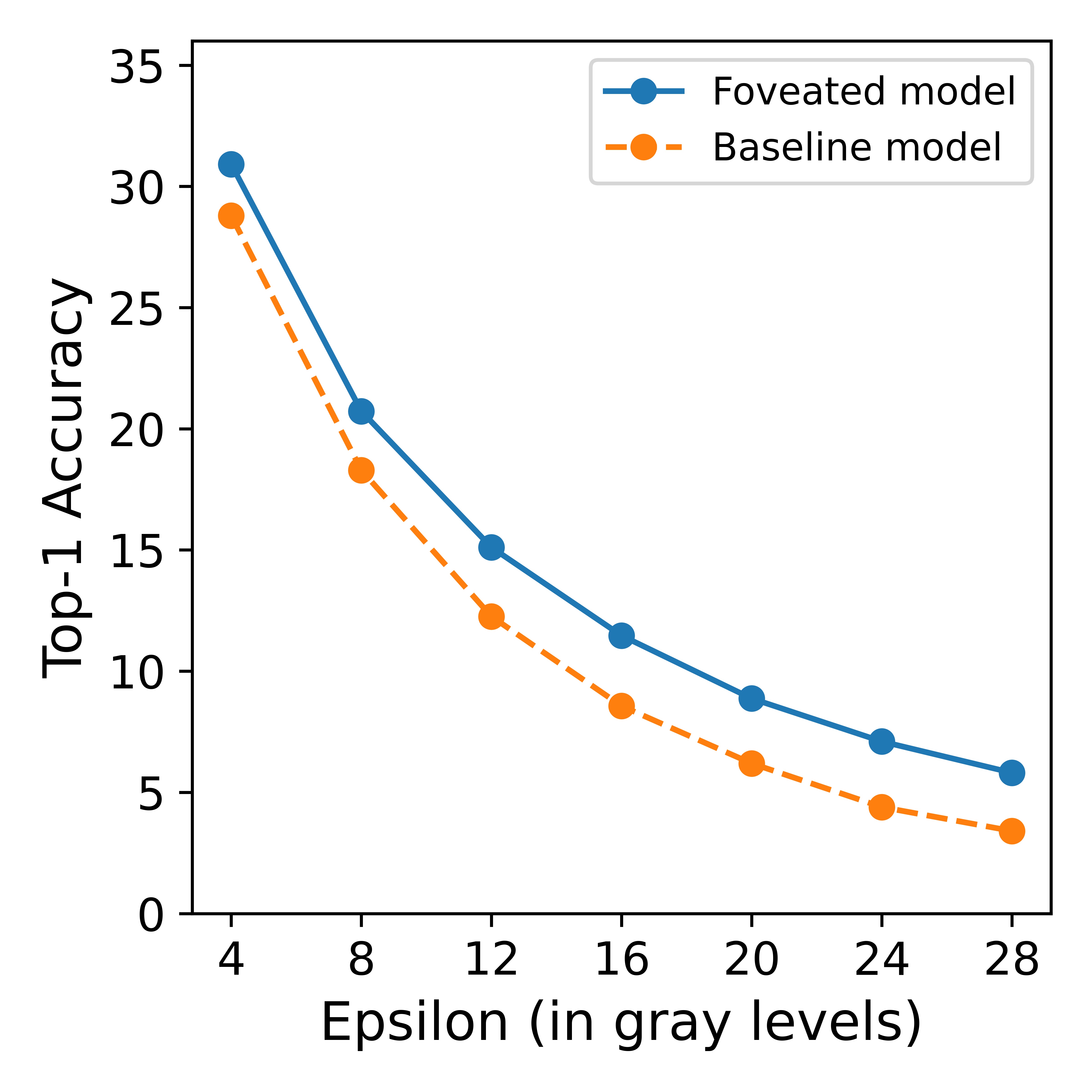}
    		\captionof{figure}{PGD attack on Foveated model with radial-polar pooling regions.}
        \label{fig:Adversarial_attack_rad_polar}
	\end{minipage}
	\vspace{-1em}
\end{table}

\section{Conclusion}

We provided a comprehensive framework for using foveal processing and fixation exploration on a Vision Transformer architecture for image classification.
The proposed architecture introduces a way to limit computations required to process an image by flexibly adjusting the required number of fixations, providing robustness to adversarial attacks, and giving us a model that can allocate computational resources based on the difficulty of an image.  Our ablation studies highlight the importance of peripheral processed features, how the self-attention guiding eye movements learn to inhibit revisits and results in accuracy similar to a model guided by predictions of human fixation (DeepGaze).  We also implemented a more biologically plausible implementation with radial polar pooling and showed that pooling parameters corresponding to visual cortical areas V1 and V4 could explain human scene categorization decisions better than the Baseline non-foveated model. 
In conclusion, we leveraged the most recent Vision Transformer architecture and combined it with ideas from foveated vision to come up with a model which has multiple knobs in terms of the number of fixations to be executed and limits on the computations performed so that the end-user will have the flexibility to fine-tune depending on their needs.

\newpage
\bibliography{iclr2023_conference}
\bibliographystyle{iclr2023_conference}

\newpage
\appendix
\section{Appendix}

\subsection{Alternate model}

We present a less biologically plausible Foveated model in this section.
With this architecture, the ensemble model can outperform the Baseline model.
 
\begin{figure}
     \centering
     \begin{subfigure}[b]{0.8\textwidth}
         \centering
         \includegraphics[width=\textwidth]{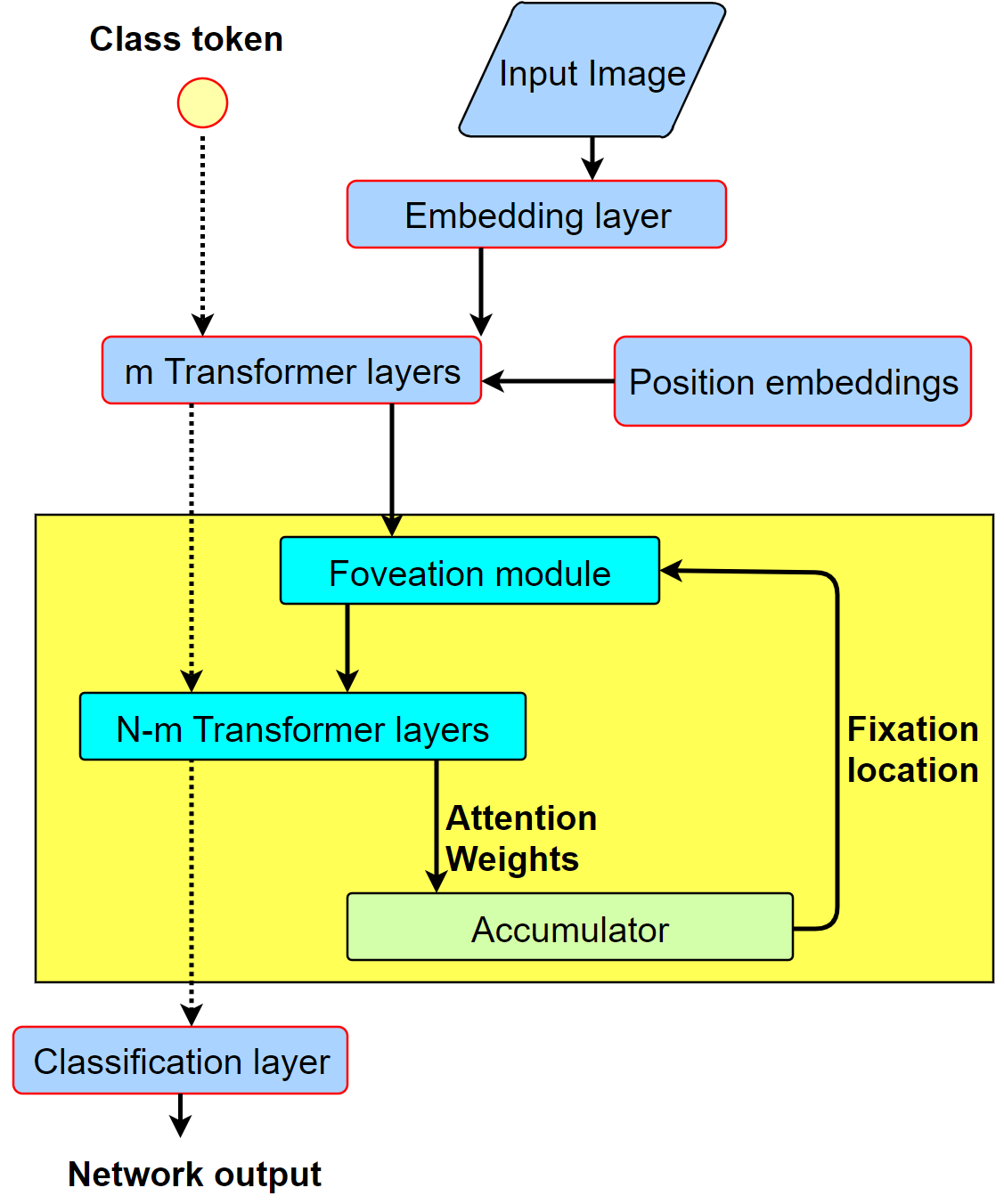}
         \caption{\noindent \textbf{FoveaTer architecture: }Solid black arrows denote the flow of image-related features. N is the total number of transformer layers. The foveation module performs fixation-dependent pooling. \textit{Accumulator} uses the attention weights from the last transformer layer of past and present fixations to predict the next fixation location. Model blocks within the yellow region are executed for each fixation.}
         \label{fig:Network_architecture}
     \end{subfigure}
     \caption{}
\end{figure}

\textbf{DeiT-Small~\cite{touvron2020deit}: } 
The DeiT-Small architecture begins with a convolution embedding layer that transforms the $[3, 224, 224]$ input image into a $[384, 14, 14]$ representation whose spacial size is $14\times14$, followed by a series of twelve transformer blocks, each sized for a 384-dimensional embedding.

\textbf{Foveated model:}
Model architecture is shown in Figure~\ref{fig:Network_architecture}.
The Foveation module can be plugged-in at any stage of the transformer architecture. 
The first \textit{m} transformer layers process full-resolution features, and the last \textit{(N-m)} transformer layers process the pooled features from the foveation module.
The input image is first passed through the embedding layer resulting in a feature vector of size $[384, 14, 14]$.
After adding the position embedding and flattening the spatial size of the embedding layer output, the resultant full-resolution feature vector of size $[384, 196]$ is passed through the \textit{m} transformer blocks along with a learnable vector of size $384$ values, called a class token.
As the same size is maintained at the input and output of the transformer layer, a feature vector of size $[384, 196]$ is obtained at the input of the Foveation module.
Then, we perform fixation-dependent average-pooling using the Foveation module,  resulting in features of size $[384, 22]$.
Under this non-uniform average-pooling model, locations closer to the fixation location use smaller neighborhoods for pooling than locations far from the fixation location.
Pooled features of size $[384, 22]$ along with the class token are passed through the remaining \textit{(N-m)} transformer layers.
We use the self-attention weights corresponding to the class token from the last transformer layer to predict the next fixation location.
Finally, the classification layer transforms the class token into a logits vector.
During training, the total number of fixations is limited to five fixations.

\begin{table}
  \caption{Throughput and Accuracy on ImageNet: We compare our models against the baseline model using Top-1 accuracy and Image throughput. (\textit{DS} - Dynamic stop, \textit{Ens} - Ensemble, \textit{Pool} - uniform $5\times5$ pooling, \textit{CF} - central fixation)}
  \label{ImageNet_Results_table}
  \centering
  \begin{tabular}{c|c|cc|cc}
    \toprule
    \multicolumn{1}{c}{Condition} & \multicolumn{1}{c}{Model} & \multicolumn{1}{c}{Fixations} & \multicolumn{1}{c}{Type} & \multicolumn{1}{c}{Throughput} & \multicolumn{1}{c}{Acc@1} \\
    \midrule
    \multirow{1}{4em}{Baseline} & DeiT-Small   &         &  Baseline  &     $323$    &   $79.83$           \\
    \midrule
     \multirow{2}{4em}{Pooled} &  DeiT-Small &         &  Pool  &    $592$     &    $73.64$  \\
     & Model &    CF-1     &     &     $564$    &    $75.2$         \\
    \midrule
    \multirow{2}{4em}{Upper bound}   & Oracle    & CF-3      &  DS   &         $507$ & $80.36$         \\
       & Oracle     & CF-3      &  Ens   &  $388$       & $84.27$            \\
    \cmidrule{2-6} 
    \multirow{2}{4em}{Optimal}   &       & CF-3         &  DS         &   $\mathbf{489}$           &  $78.31$       \\
       &      & CF-3         &  Ens   & $\mathbf{348}$    &  $\mathbf{79.99}$         \\
    \bottomrule
  \end{tabular}
\end{table}

We use a 6-6 configuration, i.e., six transformer layers before the Foveation module and six transformer layers after it.
We present the results on the ImageNet dataset in Table~\ref{ImageNet_Results_table}.
The original full-resolution model is referred to as 'Baseline', which has a throughput of $323$ and Top-1 accuracy of $79.83$. 
Since the first level of the pooling region is of size $5\times5$, we construct a pooled version of the baseline model using $5\times5$ average-pooling.
We compare this with the foveated model with one fixation at the image center, with approximately the same throughput.
The foveated model with single fixation outperforms the pooled baseline model, as shown in row $3$.
'Oracle' refers to the model with perfect Dynamic-stop, i.e., it knows the ground truth and stops the model when the prediction matches the ground truth.
Since 'Oracle' has the perfect stopping rule, it provides the upper bound on the performance of the Dynamic-stop model.
Dynamic-stop and Ensemble performance is computed.
Finally, the Foveated model's ensemble model outperforms the Baseline model in terms of throughput and accuracy.

\newpage
\subsection{Scene categories used for Psychophysics experiment}
Classes present in the scene classification task,

\begin{enumerate}
\item airport terminal
\item amphitheater
\item assembly line
\item bamboo forest
\item banquet hall
\item basement
\item beach
\item boxing ring
\item bus interior
\item canal natural
\item canyon
\item classroom
\item cliff
\item corn field
\item department store
\item desert sand
\item dining room
\item forest path
\item glacier
\item greenhouse indoor
\item gymnasium indoor
\item jail cell
\item museum indoor
\item phone booth
\item railroad track
\item sauna
\item subway station platform
\item water park
\item wind farm
\item zen garden
\end{enumerate} 

\newpage
\subsection{Comparison of FoveaTer with existing models}
\label{comparison_with_existing_models}

\begin{table}[h!]
\centering
          \begin{tabular}{c|c|c|c}
            \toprule
             & \textbf{Luo (2016)} & \textbf{Reddy (2020)} & \textbf{Ours}  \\
            \midrule

            Dataset & ImageNet & CIFAR10, ImageNet & ImageNet, Places365 subset \\
            Baseline Architecture & CNN (AlexNet, VGG, GNT) & CNN (ResNet) & Vision Transformer (deit) \\
            Image scaling & Yes & No & No \\
            Adversarial attacks & BFGS, sign method & FGSM, PGD & PGD\\
            Resource usage (N fix) & 1x & Retinal - Nx, Cortical - 1x & ~0.8x for 3 fix \\
            Foveation Location & Input image & Input image & can plug-in anywhere \\
            \bottomrule
          \end{tabular}
        
     \vspace{1em}
     \caption{Comparison with existing models}
     \label{tab:comparison_with_existing_models}
\end{table}

Comparison with existing models, which show the robustness of the foveated systems against adversarial attacks, is demonstrated in Table~\ref{tab:comparison_with_existing_models}.
Our model is based on Vision transformer architecture compared to the other models on CNN architectures. Our model can also be extended to have a convolution backbone, as shown in the supplementary material.
We do not perform any image scaling.
Our resource usage is $0.8\times$ that of the full resolution model.
We allow the possibility of applying foveation to an intermediate feature map rather than restricting it to be applied only to the input image.

\end{document}